\documentclass[10pt, a4paper]{article}
\usepackage{lrec}
\usepackage[dvipsnames]{xcolor}
\usepackage{booktabs}
\usepackage{subcaption}
\usepackage{soul}
\usepackage[normalem]{ulem}

\title{MuST-Cinema: a Speech-to-Subtitles corpus}

\name{Alina Karakanta$^{1,2}$, Matteo Negri$^1$, Marco Turchi$^1$}

\address{$^1$ Fondazione Bruno Kessler, Via Sommarive 18, Povo, Trento - Italy  \\ $^2$ University of Trento, Italy\\
        {\tt \{akarakanta,negri,turchi\}@fbk.eu}}

\abstract{Growing needs in localising audiovisual content in multiple languages through subtitles call for the development of automatic solutions for human subtitling. Neural Machine Translation (NMT) can contribute to the automatisation of subtitling, facilitating the work of human subtitlers and reducing turn-around times and related costs. NMT requires high-quality, large, task-specific training data. The existing subtitling corpora, however, are missing both alignments to the source language audio and important information about subtitle breaks. This poses a significant limitation for developing efficient automatic approaches for subtitling, since the length and form of a subtitle directly depends on the duration of the utterance. In this work, we present MuST-Cinema, a multilingual speech translation corpus built from TED subtitles. The corpus is comprised of (\textit{audio}, \textit{transcription}, \textit{translation}) triplets. Subtitle breaks are preserved by inserting special symbols. We show that the corpus can be used to build models that efficiently segment sentences into subtitles and propose a method for annotating existing subtitling corpora with subtitle breaks, conforming to the constraint of length.
 \\ \newline  }

\begin{document}

\maketitleabstract

\section{Introduction}
In the past few years, the audiovisual sector has witnessed an unprecedented growth, mostly due to the immense amounts of videos on-demand becoming available. In order to reach global audiences, audiovisual content providers localise their content into the language of the target audience. This has led to a ``subtitling boom'', since there is a huge need for offering high-quality subtitles into dozens of languages in a short time. 
However, the workflows for subtitling are complex; translation is only one step in a long pipeline consisting of transcription, timing (also called spotting) and editing. Subtitling currently heavily relies on human work and hence manual approaches are laborious and costly. 
Therefore, there is ample ground for developing automatic solutions for efficiently providing subtitles in multiple languages, reducing human workload and the overall costs of spreading audiovisual content across cultures via subtitling.

Recent developments in Neural Machine Translation (NMT) have opened up possibilities for processing inputs other than text  within a single model component. This is particularly relevant for subtitling, where the translation depends not only on the source text, but also on acoustic and visual information. 
For example, in Multimodal NMT  \cite{barrault-etal-2018-findings} 
the input can be both text and image, and Spoken Language NMT 
directly receives audio as input \cite{niehues-2019-IWSLT}. These developments can be leveraged in order to reduce the effort involved in subtitling. 

Training NMT systems, however, requires large amounts of high-quality parallel data, representative of the task. A recent study \cite{karakanta2019subtitling} questioned the conformity of existing subtitling corpora to subtitling constraints. The authors suggested that subtitling corpora are insufficient for developing end-to-end NMT solutions 
for at least two reasons; first, because they do not provide access to the source language audio or to information about the duration of a spoken utterance, and second, 
because line breaks between subtitles were removed during the corpus compilation process in order to create parallel sentences. 
Given that the length and duration of a subtitle on the screen is directly related to the duration of the utterance, the missing audio alignments pose a problem for translating the source text into a proper subtitle. Moreover, the lack of information about subtitle breaks means that splitting the translated sentences into subtitles has to be performed as part of a post-editing/post-processing step, increasing the human effort involved. 

In this work, we address these limitations
by presenting MuST-Cinema, a multilingual speech translation corpus where subtitle breaks have been automatically annotated with special symbols. The corpus is unique of its kind, since, compared to other subtitling corpora, it provides access to the source language audio, which is indispensable for automatically modelling spatial and temporal subtitling constraints.\footnote{The corpus, the trained models described in the experiments as well as the evaluation scripts can be accessed at \url{https://ict.fbk.eu/must-cinema}} 

Our contributions can be summarised as follows:
\begin{itemize}
    \item We release MuST-Cinema, a multilingual dataset with 
    (\textit{audio}, \textit{transcription}, \textit{translation})
    triplets annotated with subtitle breaks;
    \item we test the usability of the corpus to train models that automatically segment a full sentence into a sequence of subtitles;
    \item we propose an iterative method for annotating subtitling corpora with subtitling breaks, respecting the constraint of length.
\end{itemize}

\section{Subtitling}\label{sec:subtitling}
Subtitling is part of Audiovisual Translation 
and it consists in creating a short text that appears usually at the bottom of the screen, based on the speech/dialogue in a video. Subtitles can be provided in the same language as the video, in which case the process is called intralingual subtitling or \textit{captioning}, or in a different language (interlingual subtitling). Another case of intralingual subtitling is Subtitling for the Deaf and the Hard-of-hearing (SDH), which also includes acoustic information. In this paper, we refer to subtitles as the interlingual subtitles.

Subtitles are a means for facilitating comprehension and should not distract the viewer from the action on the screen. Therefore, they should conform to specific spatial and temporal constraints \cite{Avt2007sub}:
\begin{enumerate}
    \item \textbf{Length}: a subtitle should not be longer than a specific number of characters per line (CPL), e.g. 42 characters for Latin scripts, 14 for Japanese, 12-16 for Chinese.
    \item \textbf{Screen space}: a subtitle should not occupy more than 10\% of the space on the screen, therefore only 2 lines are allowed per subtitle block.
    \item \textbf{Reading speed}: a subtitle should appear on the screen at a comfortable speed for the viewer, neither too short, nor too long. A suggested optimal reading speed is 21 characters per second (CPS).
    \item \textbf{‘Linguistic wholes’}: semantic/syntactic units should remain in the same subtitle.
    \item \textbf{Equal length of lines}: the length of the two lines should be equal in order to alleviate the need of long saccades for the viewers' eyes (aesthetic constraint). 
\end{enumerate}

\begin{figure}
  \begin{subfigure}[b]{\columnwidth} 
    \includegraphics[width=\linewidth]{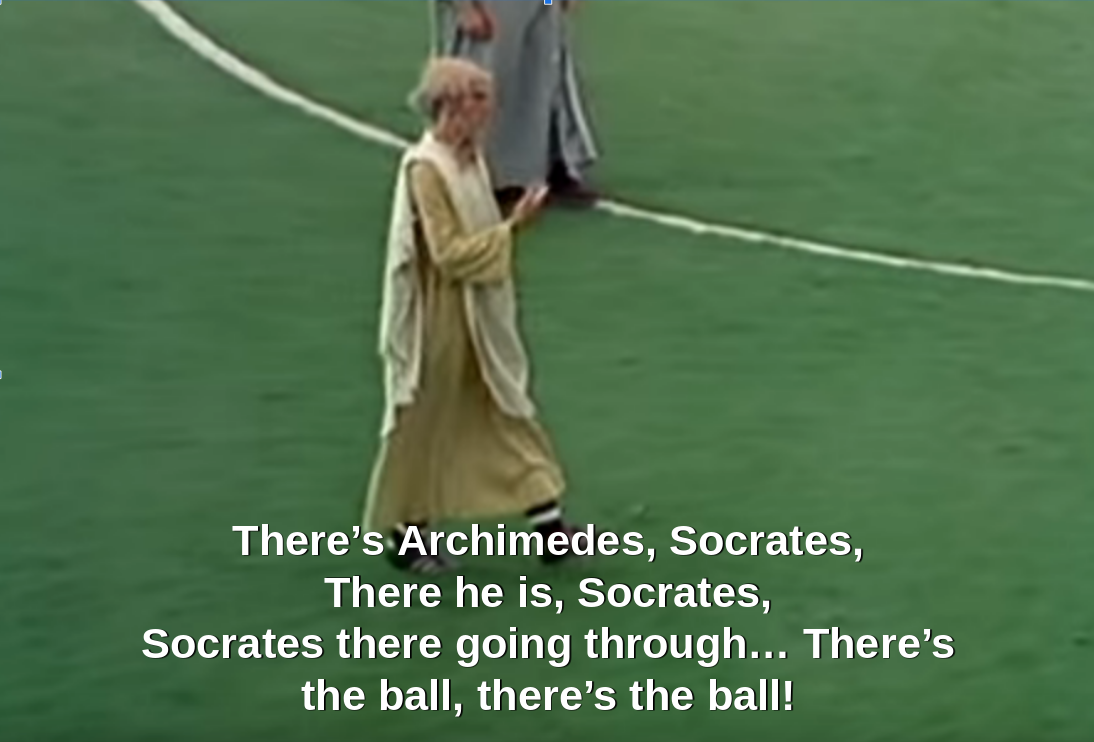}
    \caption{Original subtitle}
    \label{fig:1}
  \end{subfigure}\\
\par  \medskip \vfill
  \begin{subfigure}[b]{\columnwidth}
    \includegraphics[width=\linewidth]{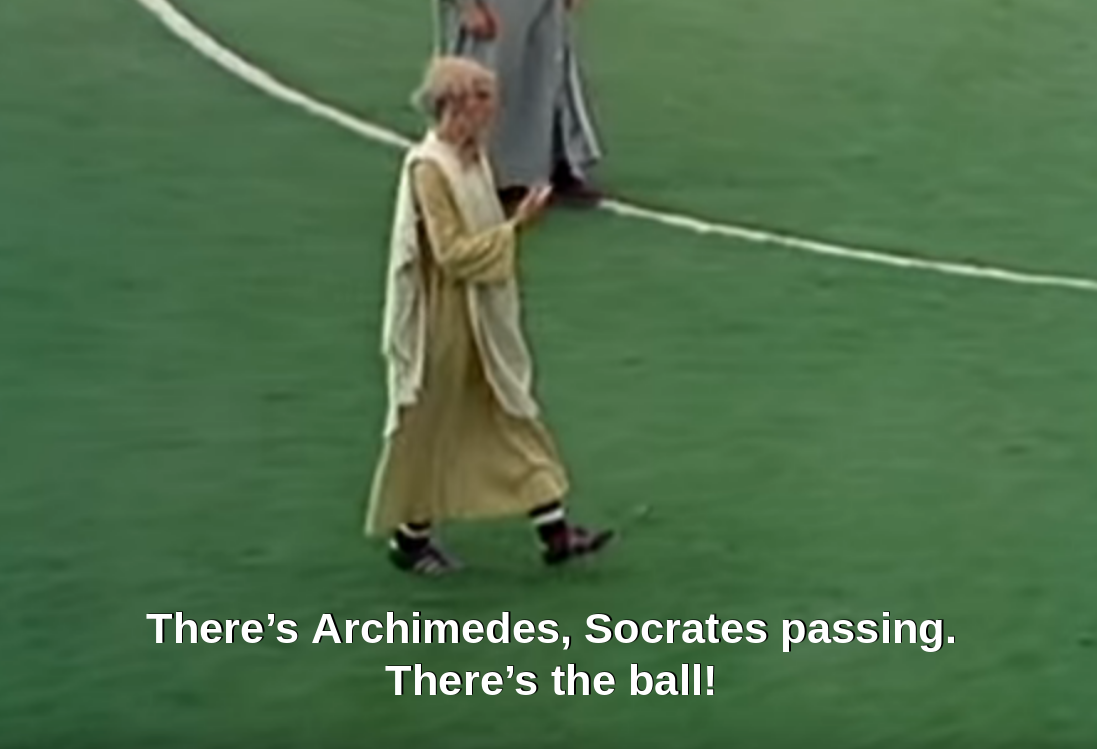}
   \caption{Subtitle adapted based on the subtitling constraints}
   \label{fig:2}
   \vfill \par  \medskip 
 \end{subfigure}
 \caption{Example of a subtitle not conforming to the subtitling constraints and the same subtitle as it should be ideally displayed.}
 \label{fig:example}
\end{figure}

Figure~\ref{fig:example} shows an example of a subtitle that does not conform to the constraints (top)\footnote{Screenshot taken from: \url{https://www.youtube.com/watch?v=i21OJ8SkBMQ}} and the same subtitle as it should be ideally displayed on the screen (bottom). 
As it can be seen, the original subtitle (Figure~\ref{fig:1}) is spread across four lines, covering almost 1/3 of the screen. Furthermore, the last two lines of the original subtitle 
are not split in a way such that  linguistic wholes are preserved: ``\textit{There's the ball}'' 
should in fact be displayed in a single line. In the bottom 
subtitle (Figure~\ref{fig:2}) instead, the lines have been kept at two, by removing redundant information and unnecessary repetitions (\textit{``Socrates, There's the ball''}). 
Each subtitle line consists of a full sentence, therefore logical completion is accomplished in each line and linguistic wholes are preserved. Lastly, \textit{``going through''} has been substituted with a synonym (\textit{``passing''}) in order not to exceed the 42-character limit. This all shows that the task of subtitlers is not limited to simply translating the speech on the screen, but they are required to compress and adapt their translation to match the temporal, spatial and aesthetic constraints. Translating subtitles can hence be seen as a complex optimisation process, where one needs to find an optimal translation based on parameters beyond semantic equivalence.

Subtitlers normally translate aided by special software that notifies them whenever their translation does not comply to the aforementioned constraints. Amara,\footnote{\url{https://amara.org/en/subtitling-platform/}} for instance, is a collaborative subtitling platform, widely used both for public and enterprise projects, as well as by initiatives like TED Talks.\footnote{\url{https://www.ted.com/}} 
In order to speed up the process, another common practice is to transcribe the speech and automatically translate it into another language. However, this process requires complex pre-and post-processing to restore the MT output in subtitle format and does not necessarily reduce the effort of the subtitler, since post-editing consists both in correcting the translation and adapting the text to match the subtitle format. 
Therefore, any automatic method that provides a translation adapted to the subtitle format would greatly simplify the work of subtitlers, significantly speeding up the process and cutting down related costs. 

\section{Related work}
In the following sections we describe several corpora that have been used in Machine Translation (MT) research for subtitling and attempts to create efficient subtitling MT systems.

\subsection{Subtitling Corpora}
The increasing availability of subtitles in multiple languages has led to several attempts of compiling parallel corpora from subtitles. The largest subtitling corpus is OpenSubtitles \cite{Lison2016OpenSubtitles2016EL}, which is built from freely available subtitles\footnote{\url{http://www.opensubtitles.org/}} in 60 languages. The subtitles come from different sources, hence converting, normalising and splitting the subtitles into sentences was a major task. The corpus contains both professional and amateur subtitles, therefore the quality of the translations can vary
. A challenge related to creating such a large subtitle corpus is the parallel sentence alignment. 
In order to create parallel sentences, the subtitles are merged and information about the subtitle breaks is removed. Even though the monolingual data, offered in XML format, preserve information about the subtitle breaks and utterance duration, mapping this information back to the parallel corpora is not straightforward. 
Another limitation is that the audiovisual material from which the subtitles are obtained is generally protected by copyright, therefore the access to audio/video is restricted, if not impossible.

A similar procedure was followed for the Japanese-English Subtitle corpus (JESC)~\cite{pryzant-jesc-2018}, a subtitle corpus consisting of 3.2 million subtitles. 
It was compiled by crawling the web for subtitles, standardising and aligning them with automatic methods. The difference with OpenSubtitles is that JESC is aligned at the level of captions and not sentences, which makes it closer to the subtitling format. Despite this, the integrity of the sentences is harmed since only subtitles with matching timestamps are included in the corpus, making it impossible to take advantage of a larger context.

Apart from films and TV series, another source for obtaining multilingual subtitles is TED Talks. TED has been hosting talks (mostly in English) from different speakers and on different topics since 2007. The talks are transcribed and then translated by volunteers into more than 100 languages, and they are submitted to multiple reviewing and approval steps to ensure their quality. 
Therefore TED Talks provide an excellent source for creating multilingual corpora on a large variety of topics. The Web Inventory of Transcribed and Translated Talks (WIT$^3$)~\cite{cettoloEtAl:EAMT2012} is a multilingual collection of transcriptions and translations of TED talks. 

Responding to the need for sizeable resources for training end-to-end speech translation systems, MuST-C \cite{mustc19} is to date the largest multilingual corpus for speech translation. Like WIT$^3$, it is also built from TED talks 
(published between 2007 and April 2019).
It contains 
(\textit{audio},  \textit{transcription},  \textit{translation}) triplets, aligned at sentence level. Based on MuST-C, the International Workshop on Spoken Language Translation (IWSLT)~\cite{niehues-2019-IWSLT}   
has been releasing data for its campaigns on the task of Spoken Language Translation (SLT).

MuST-C is a promising corpus for building end-to-end systems which translate from an audio stream directly into subtitles. However, as in OpenSubtitles, the subtitles were merged to create full sentences and the information about the subtitle breaks was removed. In this work, we attempt to overcome this limitation by annotating MuST-C with the missing subtitle breaks in order to provide MuST-Cinema, the largest subtitling corpus available aligned to the audio.

\subsection{Machine Translation for Subtitles} 
The majority of works on subtitling MT stem from the era of Statistical Machine Translation (SMT), mostly in relation to large-scale production projects. \newcite{Volk-et-al-2010} built SMT systems for translating TV subtitles for Danish, English, Norwegian and Swedish. The SUMAT project, an EU-funded project which ran from 2011 to 2014, aimed at offering an online service for MT subtitling. The scope was to collect subtitling resources, build and evaluate viable SMT solutions for the subtitling industry in nine language pairs, but there are only limited project findings available~\cite{bywood-2013-sumat,bywood-2017-sumat}. The systems involved in the above mentioned initiatives were built with proprietary data, which accentuates the need for offering freely-available subtitling resources to promote research in this direction. Still using SMT approaches, \newcite{aziz-etal_EAMT:2012} modeled temporal and spatial constraints as part of the generation process in order to compress the subtitles only in the language pair English into Portuguese.

The only approach utilising NMT for translating subtitles is described in \newcite{matusov-etal-2019-customizing} for the language pair English into Spanish. The authors proposed a complex pipeline of several elements to customise NMT to subtitle translation. Among those is a subtitle segmentation algorithm that predicts the end of a subtitle line using a recurrent neural network learned from human segmentation decisions, combined with subtitle length constraints established in the subtitling industry. Although they showed reductions in post-editing effort, it is not clear whether the improvements come from the segmentation algorithm or from fine-tuning the system to a domain which is very close to the test set.

\section{Corpus creation}
MuST-Cinema is built on top of MuST-C, by annotating the transcription and the translation with special symbols, which mark the breaks between subtitles. 
The following sections describe the process of creating MuST-Cinema for 7 languages: Dutch, French, German, Italian, Portuguese, Romanian and Spanish.

\subsection{Mapping sentences to subtitles}
As described in~\cite{mustc19}, the MuST-C corpus contains audios, transcriptions and translations obtained by processing the TED videos, and the source and target language SubRip subtitle files (.srt). 
This process consists in concatenating the textual parts of all the .srt files for each language, splitting the resulting text according to strong punctuation and then aligning it between the languages. The source sentences are then aligned to the audio by using the Gentle software.\footnote{\url{github.com/lowerquality/gentle}} Although this sequence of steps allows the authors to release a parallel corpus aligned at sentence level, the source and target subtitle properties mentioned in Section~\ref{sec:subtitling} are not preserved.  
To overcome this limitation of MuST-C and create MuST-Cinema, the following procedure was implemented in order to segment the MuST-C sentences at subtitle level:
 
 \begin{itemize}
 \item All the subtitles in the original .srt files obtained from the ted2srt website\footnote{\url{https://ted2srt.org/}} are loaded in an inverted index (text, talk ID and timestamps);
 \item For each sentence in MuST-C, the index is queried to retrieve all the subtitles that: i) belong to the same TED talk of the sentence query and ii) are fully contained in the sentence query; 
 \item An identical version of the query is created by concatenating the retrieved subtitles, and by adding special symbols to indicate the breaks.
 \end{itemize}


\subsection{Inserting breaks}
When reconstructing the sentences from the .srt files, we insert special symbols in order to mark the line and block breaks. We distinguish between two types of breaks: 1) block breaks, i.e. breaks denoting the end of the subtitle on the current screen and 2) line breaks, i.e. breaks between two consecutive lines (wherever 2 lines are present) inside the same subtitle block. 
We use the following symbols: 1) \textit{<eob>}, to mark the end of block, and 2) \textit{<eol>}, to mark the end of line.
In the case of one line per subtitle block, we use <eob>.
Figure~\ref{fig:breaks} shows and example of a sentence after inserting the breaks based on the corresponding .srt file.

\begin{figure}[h!]
    \centering
    \noindent\fbox{%
    \parbox{\linewidth}{%
    164\\
    00:08:57,020 --> 00:08:58,476 164\\
    I wanted to challenge the idea\\
    
    165\\
    00:08:58,500 --> 00:09:02,060 \\
    that design is but a tool \\
    to create function and beauty.
    }
}
\noindent\fbox{%
    \parbox{\linewidth}{%
        I wanted to challenge the idea \textcolor{TealBlue}{<eob>} that design is but a tool \textcolor{Violet}{<eol>} to create function and beauty. \textcolor{TealBlue}{<eob>}
    }%
}    \caption{Subtitle file (top) and the full sentence annotated with the subtitle breaks (bottom).}
    \label{fig:breaks}
\end{figure}{}

\subsection{Development and test sets}
For the development and test sets of MuST-Cinema, our goal is to 1) offer high-quality sets, that are also manually checked against the video, 2) avoid having subtitles above the character limit or subtitles with missing breaks and 3) preserve the integrity of the talks by including all segments. 
For these reasons we created new development and test sets for MuST-Cinema based on talks which were not part of MuST-C, for which we obtained the subtitles offered through Amara.\footnote{\url{https://amara.org/en/teams/ted/videos/}} We selected the subtitle files from Amara, because, compared to ted2srt, they contain the .srt files that are actually used in the TED videos, therefore the subtitle breaks in these files are accurate. 

For the test set, we manually selected 5 talks with subtitles available in all 7 languages, which were published after April 2019, in order to avoid any overlap with the training data. Hence, we obtained a common test set for all the languages. Each language was separately and manually aligned to the English transcription using InterText~\cite{vondricka-2014-intertext} in order to obtain parallel sentences. 

The same steps were performed for the development set, with the difference that we manually selected talks for each language without the requirement that the English talk has translations for all the languages. Therefore the development sets differ for each language, but the size of the set was kept similar for all languages.

\section{Corpus structure and statistics}
\begin{figure}[h]
    \centering
    \includegraphics[scale=.19]{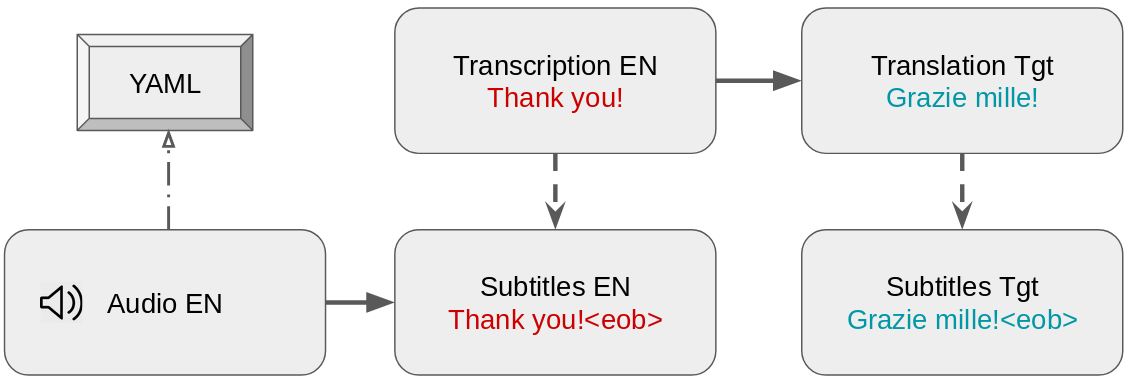}
    \caption{Structure of MuST-Cinema.}
    \label{fig:structure}
\end{figure}

The structure of MuST-Cinema is shown in Figure~\ref{fig:structure}.

There are connections at two levels; first, on the horizontal axis, the corpus is composed of triplets of audio, source language transcription annotated with subtitle breaks (\textit{Subtitles EN}), target language translation annotated with subtitle breaks (\textit{Subtitles Tgt}). Both the transcription and the translation are full sentences, aligned between English and the target language, as in MuST-C. On the vertical axis, there is a monolingual mapping between sentences with and without subtitle breaks: the EN transcription is paired to the EN subtitles of the same sentence and the Target language (Tgt) translation is paired to the Tgt subtitles of the same sentence. The utterance duration (start and end time) of each sentence is stored as metadata in a YAML file. 
The structure of MuST-Cinema makes it optimal for several tasks related to subtitling, from monolingual subtitle segmentation (as we show in Section~\ref{sec:exp}) to subtitling-oriented NMT and End-to-End speech-to-subtitling NMT.

The statistics of the corpus (train, development and test sets) are presented in Table~\ref{tab:stats}.

\begin{table}[h]
    \centering
    \begin{tabular}{l|cc|cr|cr}\toprule
     & \textbf{Train} & & \textbf{Dev} & & \textbf{Test} \\ \midrule
        \textbf{Tgt} & \textbf{sents} & \textbf{tgt w} &  \textbf{sents} & \textbf{tgt w} & \textbf{sents} & \textbf{tgt w}\\ \midrule
         \textbf{De} & 229K & 4.7M & 1088 & 19K & 542 & 9.3K\\
         \textbf{Es} & 265K & 6.1M & 1095 & 21K & 536 & 10K\\
         \textbf{Fr} & 275K & 6.3M & 1079 & 22K & 544 & 10K\\
         \textbf{It} & 253K & 5.4M & 1049 & 20K & 545 & 9.6K\\
         \textbf{Nl} & 248K & 4.8M & 1023 & 20K & 548 & 10K\\
         \textbf{Pt} & 206K & 4.2M & 975 & 18K & 542 & 10K\\
         \textbf{Ro} & 236K & 4.7M & 494 & 9K & 543 & 10K\\ \bottomrule
    \end{tabular}
    \caption{Number of sentences and target side words for each language of MuST-Cinema for the training, development and test set for all 7 languages.}
    \label{tab:stats}
\end{table}

\subsection{Analysis and subtitle quality}\label{sec:corpus_ana}







MuST-Cinema incorporates missing elements of other subtitling corpora as described in \cite{karakanta2019subtitling}, by: i) providing the duration of the utterance, both through the alignment to the audio and as metadata, ii) preserving the integrity of the documents, since the talks in the dev and test sets were used as a whole and without shuffling the sentences, iii) including subtitle breaks as annotations and iv) offering a reliable gold standard for evaluation of subtitling-related tasks.
For the analysis of the corpus, we focus on the primary aspects of subtitling quality, subtitle length and number of lines. We explore this aspect in breadth, for all the 7 languages of MuST-Cinema. 



\begin{figure}[t]
    \centering
    \includegraphics[width=\columnwidth]{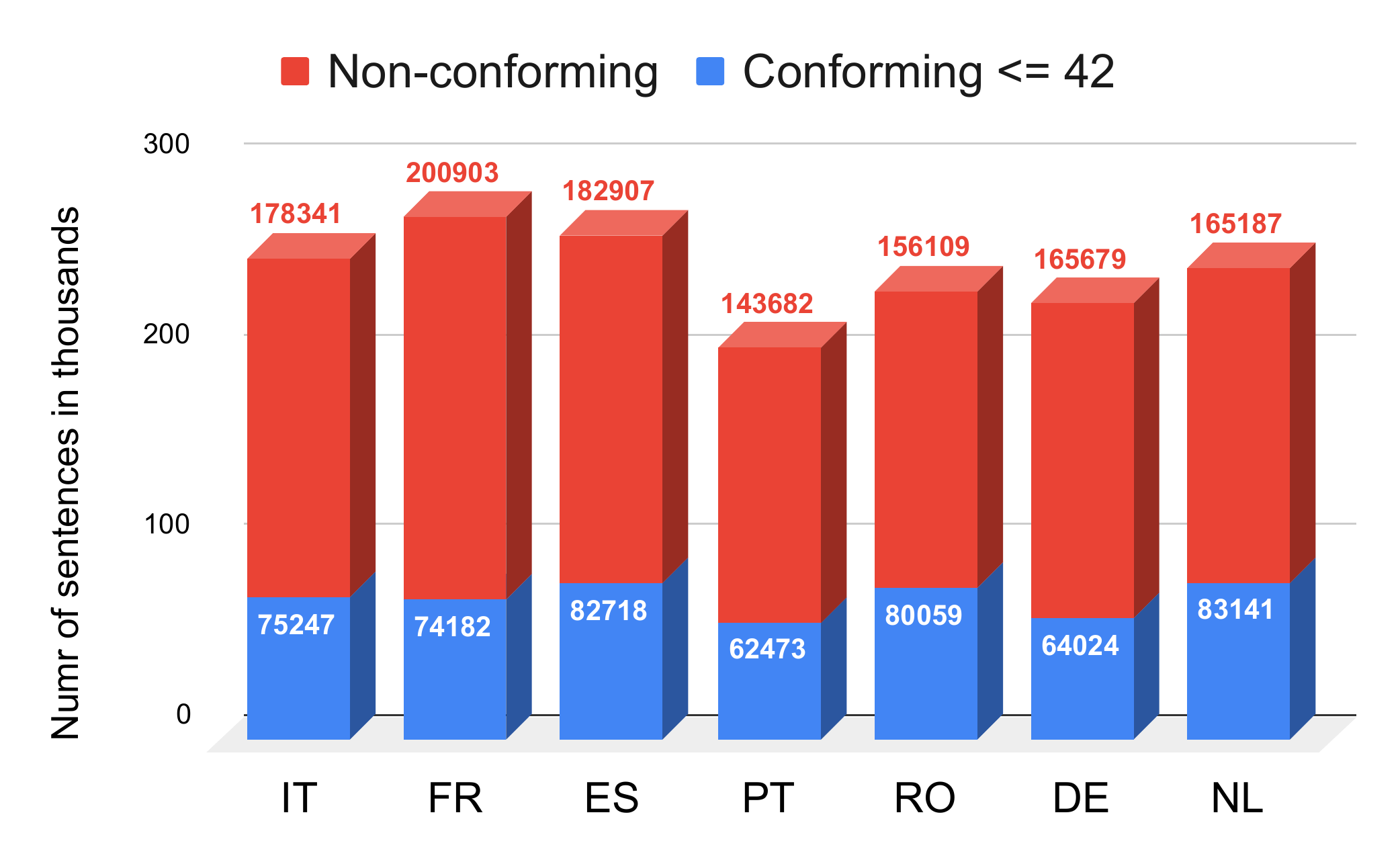}
    \caption{Statistics about the sentences conforming to the subtitling constraint of length (blue) and not-conforming (red) for CPL$<=$42.}
    \label{fig:42}
\end{figure}
\begin{figure}[h]
    \centering
    \includegraphics[width=\columnwidth]{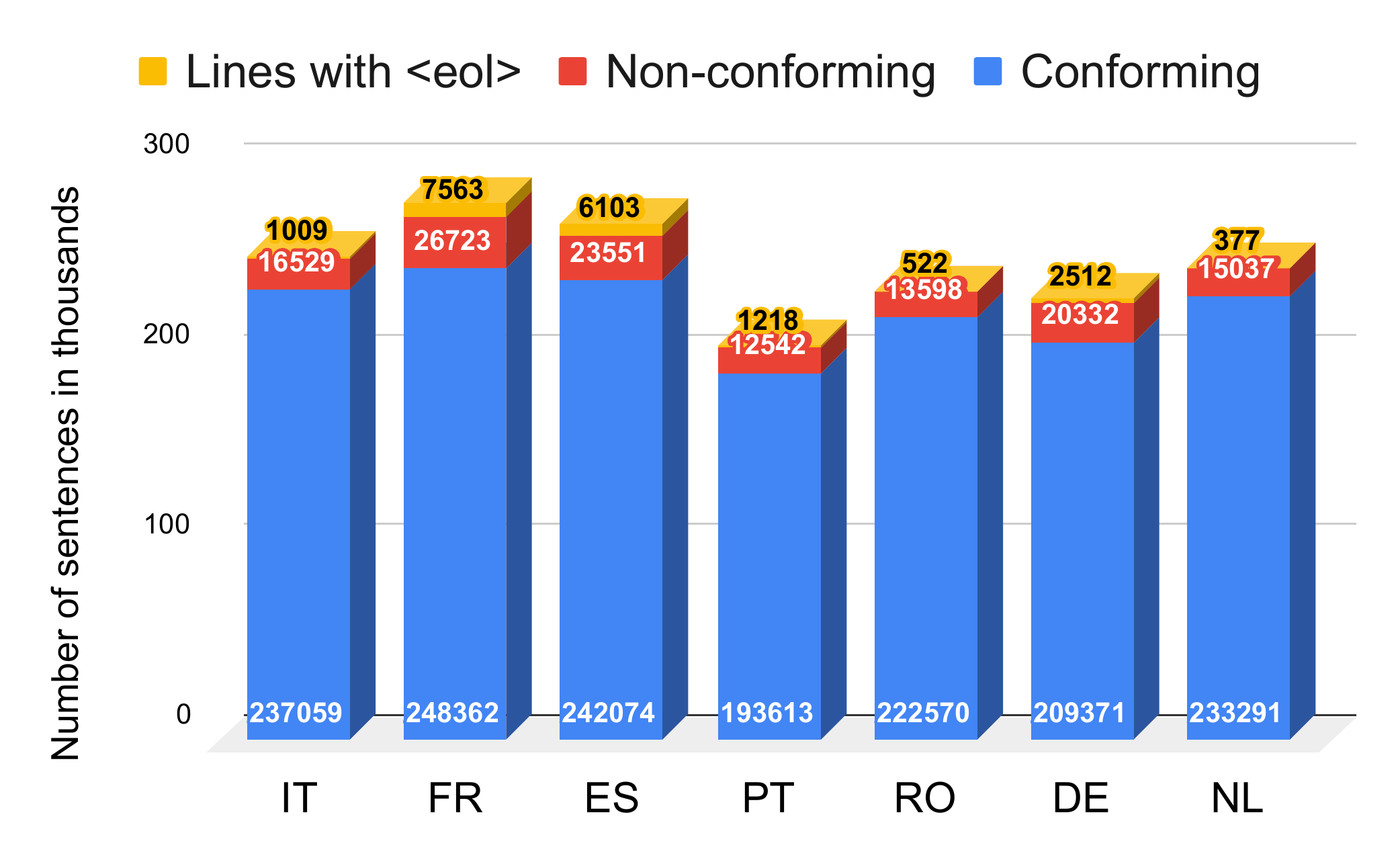}
    \caption{Statistics about the sentences conforming to the subtitling constraint of length (blue), not-conforming (red) for CPL$<=$84 and number of sentences containing information about the <eol> (yellow).}
    \label{fig:cpl}
\end{figure}

For length, we considered as conforming sentences the sentences for which each one of the subtitles composing it has a length $<=42$ characters. Understandably, this is a strict criterion, since the sentence is considered non-conforming even if just one of the subtitles composing it is longer than the 42-threshold. Figure~\ref{fig:42} shows that nearly 70\% of the sentences are non-conforming. An inspection of the data showed that the subtitle blocks in the .srt files downloaded from ted2srt are never composed of two lines. When comparing the subtitles in the .srt files with the video published on the TED website, we noticed that the line breaks inside subtitle blocks were removed from the .srt files (for reasons not explained anywhere). 
We observed though, that in some talks the subtitle blocks contained a double space, which corresponds to the <eol> in the videos. We replaced the double spaces with the <eol> symbol. We concluded that the subtitle lines inside the blocks were collapsed into one sentence. In fact, when we performed the analysis of the CPL with a maximum length of 84 ($2*42$), more than 90\% of the sentences conform to the length criterion (Figure~\ref{fig:cpl}). The top 
portions in the figure show the number of sentences where the <eol> symbol is present. This is a small portion of the data, however, later we will show that it can valuable for learning subtitle line segmentations. This is an issue present only in the training data, as the development and test sets were manually downloaded from the TED Amara website, where the subtitle lines inside the .srt files are not collapsed. In this work, instead of attempting to recreate MuST-C from the Amara website (which would require a new pipeline and weeks for downloading and processing the dumps), we re-annotate the existing data with a model that learns to insert the <eol> symbol (Section~\ref{sec:exp}).

\section{Experiments}\label{sec:exp}
In the following sections, we present the experiments for the task of monolingual segmentation of a sentence into subtitles. 

We train a \textit{segmenter}, i.e. a model that inserts subtitle breaks given a non-annotated sentence monolingually. 
Here, we address the problem of splitting the sentences into subtitles as a sequence-to-sequence problem, where the input is the full sentence and the output is the same sentence annotated with <eol> and <eob> symbols. 

\subsection{Experimental Setting}
We create two portions of training data; one containing all the parallel sentences available, 
 and a smaller one containing only the sentences where <eol> symbols are present (top parts in Figure~\ref{fig:cpl}). We segment the data into subword units with SentencePiece\footnote{\url{https://github.com/google/sentencepiece}} with a different model for each language and 8K vocabulary size. The break symbols are maintained as a single word-piece. 

We train a sequence-to-sequence model based on the Transformer architecture \cite{vaswani2017attention} using fairseq~\cite{ott2019fairseq}. The model is first trained on all the available data (\textit{All}) and then fine-tuned on the sentences containing the <eol> 
symbols (\textit{ft\_eol}). This two-step procedure was applied to force the NMT system to take into consideration also the <eol> symbols. This was necessary considering the mismatch in the number between sentences with and without <eol> symbols. For optimisation, we use Adam~\cite{kingma2014adam} (betas 0.9, 0.98), and dropout~\cite{srivastava2014dropout} is set to 0.3. We train the initial model for 12 epochs and fine-tune for 6 epochs.

As baseline, we use an algorithm based on counting characters (\textit{Count Char}). The algorithm consumes characters until the maximum length of 42 is reached and then inserts a break after the last consumed space. If the previous break is an <eob>, it randomly selects between <eob> and <eol>, while if the previous break is an <eol>, it inserts an <eob>. This choice is motivated by the fact that, due to the constraint of having only two lines per block, we cannot have a segmentation containing consecutive <eol> symbols. The last break is always an <eob>.

\subsection{Evaluation}
We evaluate our models using 3 categories of metrics.
First, we compute the BLEU~\cite{papineni2002bleu} between the human reference and the output of the system. We report BLEU score 1) on the raw output, with the breaks (general BLEU) and 2) on the output without the breaks. The first  computes the n-gram overlap between reference and output. Higher values indicate a high similarity between the system's output and the desired output.
The second ensures that no changes are performed to the actual text, since the task is only about inserting the break symbols, without changing the content. Higher values indicate less changes between the original text and the text where the break symbols were inserted. 

Then, we test how good the model is at inserting a sufficient number breaks and at the right places. Hence, we compute the precision, recall and F1-score, as follows:
\begin{equation}
    Precision = \frac{\#correct\_breaks}{\#total\_breaks(output)}
\end{equation}

\begin{equation}
    Recall = \frac{\#correct\_breaks}{\#total\_breaks(reference)}
\end{equation}

\begin{equation}
    F1 = 2 * \frac{precision*recall}{precision+recall}
\end{equation}{}

Finally, we want to test the ability of the system to segment the sentences into properly formed subtitles,~i.e. how well the system conforms to the constraint of length and number of lines. 
Therefore, we compute CPL to report the number of subtitles conforming to a length $<=42$.

\section{Results and Analysis}

\begin{figure}[h!]
  \includegraphics[width=\columnwidth]{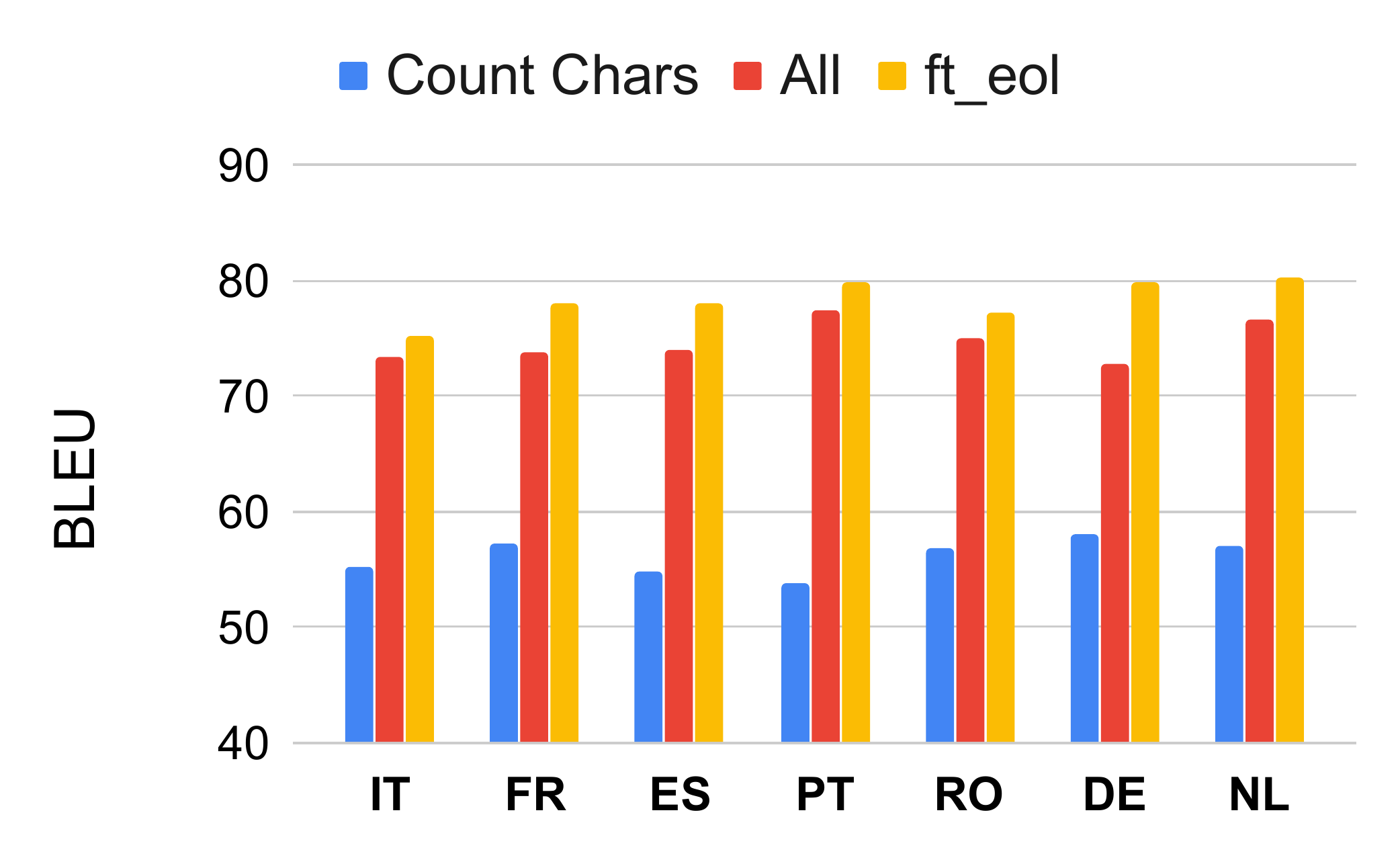}
  \caption{BLEU (general).}\label{fig:bleu}
\end{figure}

The results for the three categories of metrics are shown in the figures. 
Figure~\ref{fig:bleu} shows the general BLEU (with breaks) of the segmented output compared to the human reference. 
The count characters (baseline) algorithm performs poorly, while the sequence-to-sequence models reach a BLEU score of above 70 for all the languages, with the model fine-tuned on the <eol> achieving an extra improvement (up to 7 points for En-De). The BLEU without the breaks (not reported here) was  always above 98\%, indicating that there were not significant changes to the actual text. Upon inspection of  the few differences, we noticed that the system corrected inconsistencies in the reference, such as apostrophes (\textit{J\textquotesingle ai} to \textit{J’ai}) or removed spaces before punctuation. The BLEU score, although useful to indicate the changes compared to the reference, 
is not very informative on the type of mistakes in inserting the breaks.

\begin{figure}[h!]
  \includegraphics[width=\columnwidth]{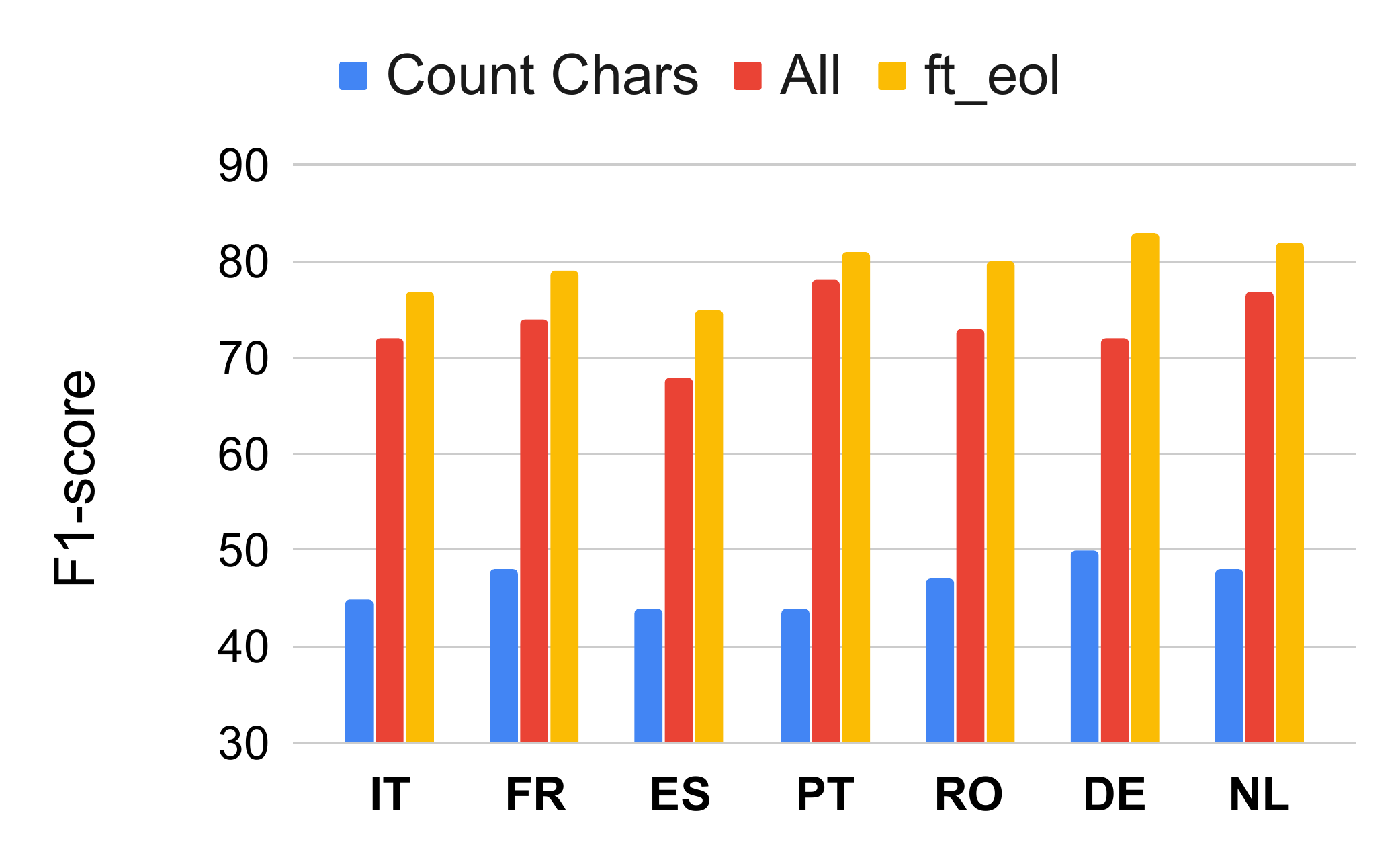}
  \caption{F1-score.}\label{fig:f1}
\end{figure}

The picture is similar for the F1-score (Figure~\ref{fig:f1}). Counting characters barely reaches 50\%, and the gap between precision and recall is small (both between 42\%-50\% for all languages). The fine-tuned model is better than the \textit{All} with 6 points on average. It is interesting noticing the trade-off between precision and recall; while precision drops with 7 points on average when fine-tuning compared to the non-fine-tuned model, recall increases by 13 points. In fact, for every fine-tuning epoch, precision drops by about 1 point compared to the previous epoch while recall increases at a double rate. This shows that fine-tuning helps in segmenting the sentence into smaller pieces by inserting more breaks, even though small differences in positioning them are highly penalised in terms of precision.
\begin{figure}[h!]
  \includegraphics[width=\columnwidth]{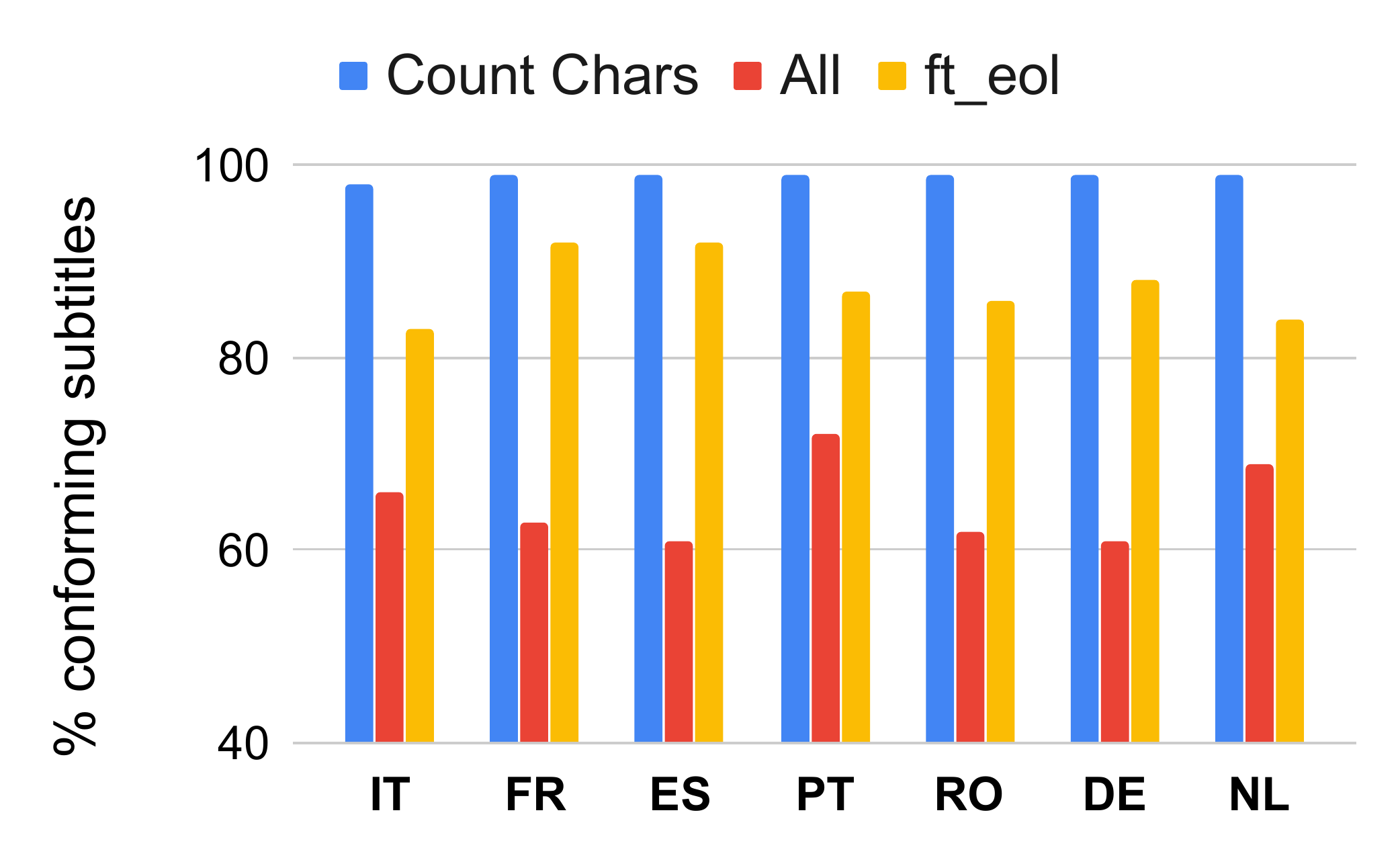}
  \caption{Percentage of subtitles conforming to the length constraint.}\label{fig:prc}
\end{figure}

Figure~\ref{fig:prc} shows the percentage of subtitles which are $<=$42 characters long. The method of counting characters conforms for almost all sentences; there is a minimum loss due to ``orfans'', which are subtitles containing less than 5 characters. Despite this, the low F1-score of the breaks  
shows that the breaks are not inserted in the correct position. Moreover, the model trained on all data (\textit{All}) does not insert sufficient breaks to reduce the length of the subtitles inside the sentence. Here the fine-tuned model achieves on average a 21\% improvement, reaching up to 92\% of length conformity for French and Spanish. In spite of the limited amount of data, the proposed fine-tuning method shows that the corpus is a valuable resource towards achieving outputs conforming to the subtitle format.

Table~\ref{tab:examples} contains some examples of the output of the segmenter:
\begin{table*}[t]
    \centering
    \begin{tabular}{|ll|}\toprule
    1. & La forma del bottone non è variata molto \textcolor{TealBlue}{<eol>} da quella che era nel Medioevo. \textcolor{TealBlue}{<eob>}\\
    & \textit{The design of a button hasn't changed much since the Middle Ages.}\\
    2. & Sie lief aus dem Haus des Mannes fort, \textcolor{TealBlue}{<eol>} den sie nicht heiraten wollte, \textcolor{TealBlue}{<eob>} \\
         & und heiratete schließlich \textcolor{TealBlue}{<eol>} den Mann ihrer Wahl. \textcolor{TealBlue}{<eob>}\\
         &  \textit{She ran away from the house of the man she did not want to marry and}\\
         & \textit{ended up marrying the man of her choice.}\\
    3. & C'est donc toujours plus difficile. \textcolor{TealBlue}{<eob>}\\
    & \textit{So it gets to be more and more of a challenge.} \\
    4. & Je m'enroule en une petite <eol> boule comme un foetus. <eob>\\
     & \textit{And I curl up into a little fetal ball.}\\
          \bottomrule
    \end{tabular}
    \caption{Examples of sentences output by the segmenter. Text in blue indicates insertions by the segmenter.}
    \label{tab:examples}
\end{table*}
Sentence 1 shows that the segmenter is able to insert breaks in the correct place. In Sentence 2, the segmenter inserts both <eol> and <eob> symbols, correctly alternating between the two. Sentence 3 is a sentence shorter than 42 characters, which was not changed by the segmenter. This shows that the segmenter doesn't insert breaks for sentences that do not need to be further split. Lastly, Sentence 4 is already annotated with breaks. In this case, the segmenter does not insert any breaks because the sentence is already segmented.

\subsection{Iterative Re-annotation}
The analysis of the corpus in Section~\ref{sec:corpus_ana} showed that the majority of the sentences in the training data of MuST-Cinema do not contain <eol> breaks because this information has been erased from the .srt files. We attempt to address this inadequacy by applying the models in Section~\ref{sec:exp} to iteratively re-annotate the training data only with the missing <eol> symbols. The difference with the models in Section~\ref{sec:exp} is that the input sentences to the segmenter already contain the <eob> symbols. This is because we wanted to preserve the subtitle block segmentation of the data while splitting the blocks into subtitle lines. The process is carried out as follows: 1) Using the best-performing models, that is the models fine-tuned on a small amount of data containing <eol> symbols (\textit{ft\_eol}), we annotate sentences in the training data which do not respect the length constraint. 2) We filter the annotated data with the CPL criterion and select the length-conforming sentences containing <eol> symbols. Moreover, we make sure that no sentences contain two or more consecutive <eol> symbols, even though we observe that this is rarely the case. 
3) 
Finally, we concatenate the selected data in step 2) with the initial data containing <eol> and fine-tune again the base model (\textit{All}). The procedure can be iterated until all sentences in the corpus conform to the criterion of length. 
Our intuition is that, even though sub-optimal due to possible insertion errors done by the system, the new training instances collected at step \textit{t} can be used to improve the model at step \textit{t+1}.

\begin{figure}[h]
    \centering
    \includegraphics[width=\columnwidth]{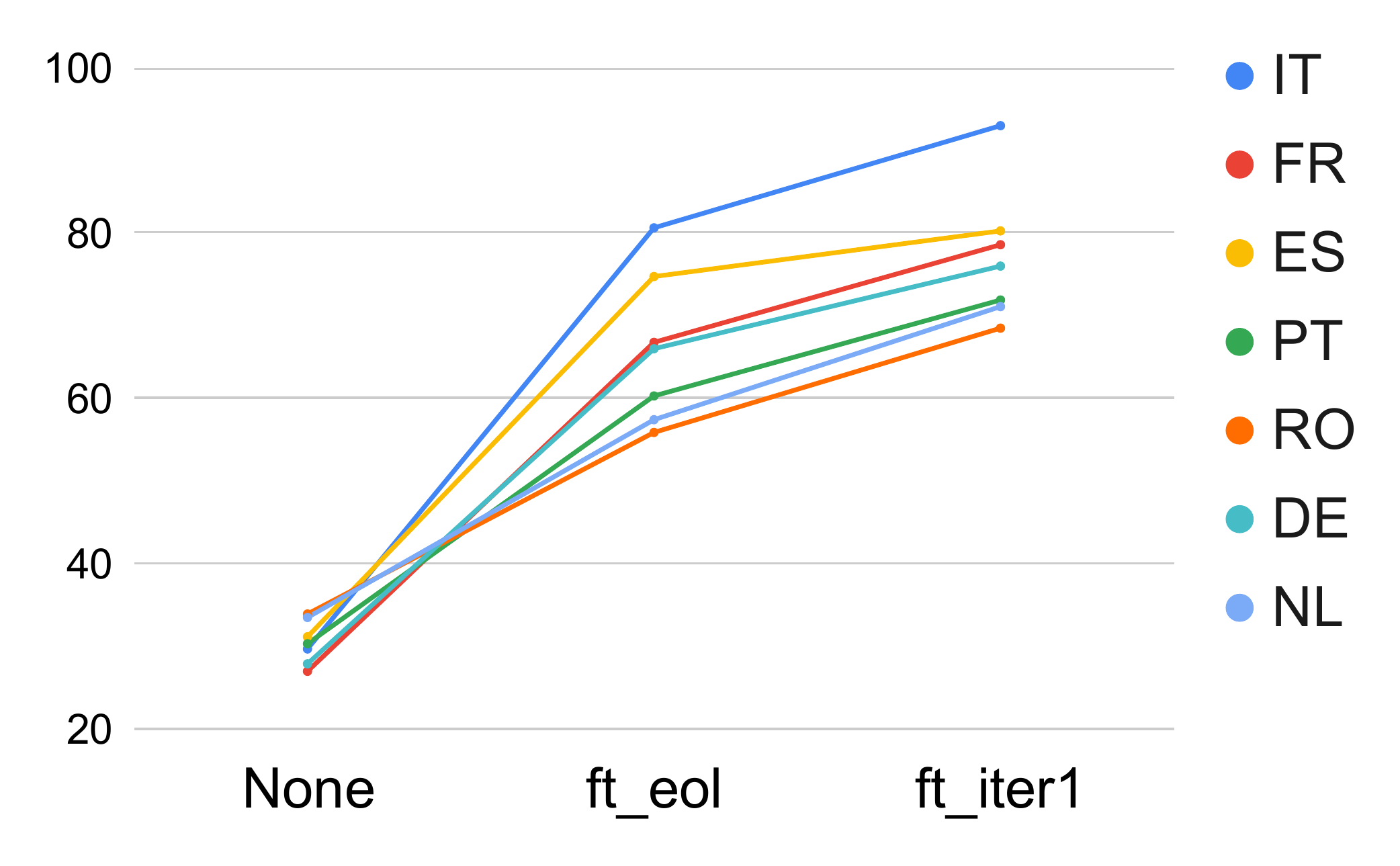}
    \caption{Percentage of subtitles in the MuST-Cinema training data conforming to CPL$<42$ after segmenting with different models.}
    \label{fig:iter}
\end{figure}

Figure~\ref{fig:iter} shows the percentage of length-conforming subtitles for the training data of MuST-Cinema at 3 stages; at the initial corpus compilation stage (\textit{None}), after segmenting the data with the model fine-tuned on small amounts of data with <eol> (\textit{ft$\_$eol}), and after 1 iteration of the method proposed above (\textit{ft$\_$iter1}). There is a clear increase in the number of length-conforming subtitles, reaching up to 92\% for Italian after only one iteration. This shows that training data can be efficiently and automatically annotated with the missing line breaks.

\section{Conclusions and future work}
There are missing elements in the existing subtitling corpora which hinder the development of automatic NMT solutions for subtitling. These elements are i) lack of audio/utterance duration, and ii) information about subtitle breaks. MuST-Cinema addresses these limitations by mapping the parallel sentences to audio and annotating the sentences with subtitle breaks.
Based on MuST-Cinema, we proposed an automatic method that can be used to annotate other subtitling corpora with subtitle breaks. Our proposed segmenter shows that full sentences can be successfully segmented into subtitles conforming to the subtitling constraints. In the future, we are planning to use the corpus for building translation models with text and audio as input. We hope MuST-Cinema will be used as a basis for kick-starting the development of end-to-end NMT solutions for subtitling. 

\section*{Acknowledgements}
This  work  is  part  of  a  project  financially  supported  by  an Amazon AWS ML Grant.

\section{Bibliographical References}
\label{main:ref}
\bibliographystyle{lrec}
\bibliography{lrec2020W-xample}

\end{document}